\documentclass[%
 amsmath,amssymb,
]{revtex4-2}

\usepackage{dcolumn} 
\usepackage{bm}      
\usepackage{setspace}
\usepackage{amsfonts}
\usepackage{braket}

\usepackage{graphicx}

\usepackage{amsmath}	
\begin{document}


\title{A thin and soft optical tactile sensor for highly sensitive object perception}

\author{
Yanchen Shen$^{1}$,
Kohei Tsuji$^{1}$,
Haruto Koizumi$^{1}$,
Jiseon Hong$^{1}$,
Tomoaki Niiyama$^{2}$,
Hiroyuki Kuwabara$^{3}$,
Hayato Ishida$^{3}$,
Jun Hiramitsu$^{3}$,
Mitsuhito Mase$^{3}$, 
and 
Satoshi Sunada$^{2}$}

\affiliation{${}^{1}$Graduate School of Natural Science and Technology, Kanazawa University, Kakuma-machi, Kanazawa, Ishikawa, 920-1192, Japan\\
${}^{1}$Faculty of Mechanical Engineering, Institute of Science and Engineering, Kanazawa University, Kakuma-machi, Kanazawa, Ishikawa, 920-1192, Japan\\
${}^{3}$Hamamatsu Photonics K.K., 1126-1, Ichino, Chuo-ku, Hamamatsu, Shizuoka, 435-8558, Japan}


\begin{abstract}
Tactile sensing is crucial in robotics and wearable devices for safe perception and interaction with the environment. Optical tactile sensors have emerged as promising solutions, as they are immune to electromagnetic interference and have high spatial resolution. However, existing optical approaches, particularly vision-based tactile sensors, rely on complex optical assemblies that involve lenses and cameras, resulting in bulky, rigid, and alignment-sensitive designs. In this study, we present a thin, compact, and soft optical tactile sensor featuring an alignment-free configuration. The soft optical sensor operates by capturing deformation-induced changes in speckle patterns generated within a soft silicone material, thereby enabling precise force measurements and texture recognition via machine learning. The experimental results show a root-mean-square error of 40 \,mN in the force measurement and a classification accuracy of 93.33\% over nine classes of textured surfaces, including Mahjong tiles. The proposed speckle-based approach provides a compact, easily fabricated, and mechanically compliant platform that bridges optical sensing with flexible shape-adaptive architectures, thereby demonstrating its potential as a novel tactile-sensing paradigm for soft robotics and wearable haptic interfaces.
\end{abstract}

\maketitle
\section{Introduction}

Mimicking the multiple mechanical sensing capabilities of the human skin has been a compelling goal in robotics and human--machine interaction. Soft tactile sensing systems aim to detect mechanical stimuli such as pressure, strain, temperature, and surface texture, enabling intelligent systems to interact safely and dexterously in complex environments~\cite{Wang2015, Sekitani2012, LEOGRANDE2025100312, MERIBOUT2024115332, Hammock2013, Jung2014, Yang2019, YAO2024111040}.

In the early 1970s, researchers began exploring tactile sensing in artificial systems, developing pioneering devices such as prosthetic hands with rudimentary feedback and touch-sensitive computer interfaces~\cite{VEDEL1982289, clippinger1974}. However, these seminal sensors were typically based on rigid materials and discrete transducers, providing limited spatial resolution and poor mechanical compliance. Over the last few decades, advances in flexible electronics and soft materials have led to significant progress in soft tactile sensors, enabling conformable skin-like devices capable of distributed and multimodal sensing \cite{Grzegorz2005, Cheng2009}.

Tactile sensors used in soft systems can be broadly categorized into four types: piezoresistive, piezoelectric, capacitive, and optical, each with distinct operating principles and trade-offs. Piezoresistive sensors convert mechanical deformations into changes in resistance, providing straightforward readout circuitry and high sensitivity. However, they often suffer from hysteresis and drift~\cite{Stassi2014,Choong2014,6584238,1641896}. Piezoelectric sensors use piezoelectric materials to generate electrical signals in response to dynamic pressure. Consequently, these sensors are suitable for vibration detection, but they are less effective for measuring static force~\cite{Qi2023, RYU2022131308}. Capacitive sensors detect changes in electrode spacing or overlap. These sensors provide excellent sensitivity and low power consumption; however, they are susceptible to electromagnetic interference and require precise fabrication~\cite{Lee2011, Lipomi2011, Zang2015}.

By contrast, optical tactile sensors offer unique advantages, including immunity to electromagnetic noise, high spatial resolution, and the potential for multiplexed sensing over long distances~\cite{YAO2024111040}. Various optical implementations have been reported, including those based on silica fibers~\cite{Mohamed2022, Kao1966, s17010111}, polymer fibers~\cite{Mizuno:21, Ma2002}, hydrogel~\cite{Guo2016, Wang2020Hydrogel, ELSHERIF201925}, micro/nanofibers (MNFs)~\cite{zhang2020mnf, Jiang2021, Pan2020}, and vision-based sensors~\cite{GelForce}. Among these, vision-based tactile sensors (VBTSs) have emerged as a promising approach~\cite{VBTSreviewAPR, VBTSreviewIEEE}. By combining a deformable elastomer with a vision system, VBTSs can reconstruct high-resolution 3D surface topographies from contact-induced deformations, as exemplified by the well-known GelSight sensors~\cite{gelsight, gelslim, gelslim3}. Moreover, numerous other studies have proposed various novel approaches, such as ChromaTouch~\cite{scharff2022rapid}, BioTacTip~\cite{BioTacTip2024}, MagicTac~\cite{magictac2024}, C-Sight~\cite{Csight2024}, and ViTacTip~\cite{vitactip2024}.

Despite their high performances, VBTS systems have several limitations. They typically require high-resolution image sensors, precise optical alignment, and controlled illumination, making them sensitive to the focus distance and environmental conditions. Moreover, its integration into compact or highly flexible platforms is challenging due to the complex optical paths and rigid camera setups.

In this study, we present a thin, compact, and highly flexible optical tactile sensor featuring an alignment-free configuration. The sensing principle is based on speckle patterns generated within a soft transparent silicone elastomer. Specifically, mechanical deformation of the elastomer is encoded into the speckle pattern via light scattering within the material~\cite{Shimadera2022, Kitagawa2024}. This optical interferometric technique enables the detection of minute optical path changes induced by subtle mechanical deformations, allowing accurate tactile sensing with a minimal number of components without requiring precise alignment. We demonstrate that the proposed tactile sensor achieves force sensing in the order of tens of millinewtons (mN), along with high-accuracy surface texture recognition, without relying on complex microstructures or intricate illumination schemes. This approach offers a simple, scalable, and geometrically versatile platform for next-generation optical tactile perception with promising applications in soft robotics and wearable systems.

\section{Concept and Operating Principle}
\begin{figure}[htbp]
    \centering
   \includegraphics[bb=0.000000 0.000000 340.000000 377.000000, width=10cm]{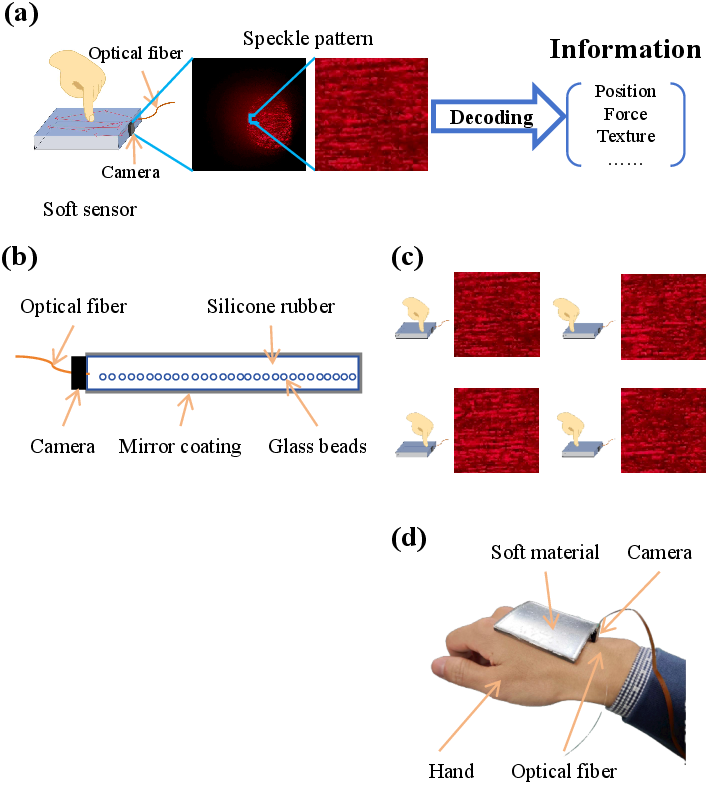}
    \caption{(a) Schematic of the sensor operation. The camera captures the speckle pattern induced by laser illumination in the soft material. A central $128 \times 128$ region is cropped and input to a decoding model (CNN) for tactile information extraction. (b) Structure of the proposed sensor. (c) Speckle patterns collected under different contact conditions. (d) Proposed sensor on a human hand.}
    \label{fig1}
\end{figure}

The sensor operation workflow is illustrated in Fig.~\ref{fig1}(a). The proposed speckle-based tactile sensor comprises a transparent soft silicone elastomer as the sensing body, an optical fiber for delivering laser light to the elastomer, and a compact image sensor (e.g., a CMOS camera). Figure~\ref{fig1}(b) shows the structure of the proposed sensor. Upon external contact, the mechanical deformation of the elastomer changes the position of the embedded scattering centers (e.g., glass microspheres), thereby modulating the internal light propagation paths. This modulation induces a reproducible transformation of the laser speckle pattern captured by the camera. Figure~\ref{fig1}(c) illustrates representative speckle patterns collected under different contact conditions. A fixed area is cropped from the image, and the encoded tactile information is decoded from the cropped image using a decoding model.

Owing to the high refractive-index contrast between silicone and air, light rays striking the interface beyond the critical angle undergo total internal reflection, thereby confining most of the light within the elastomer. Reflective coatings are applied to the surfaces to enhance light confinement and sensing sensitivity, ensuring that light undergoes multiple passes through the scattering region. This extended propagation amplifies the interaction between light and mechanical deformation, increasing both the effective scattering depth and sensitivity. 

The core transduction mechanism of the proposed sensor is based on encoding tactile information through optical scattering within the material. This mechanism is fundamentally different from those used in conventional VBTSs, which typically rely on surface markers or structured illumination. The optical scattering-based approach offers several advantages: (i) high-precision and multimodal sensing enabled by its interferometric characteristics; (ii) elimination of the need for focusing lenses and precise optical alignment; (iii) reduced dependence on camera pixel resolution, enabling high-performance tactile sensing even with compact, low-resolution imaging hardware; and (iv) high geometric flexibility of the sensor architecture, allowing conformity to arbitrarily curved surfaces. For example, as shown in Fig.~\ref{fig1}(d), the proposed sensor can be comfortably mounted on a human hand, demonstrating its adaptability to complex nonplanar surfaces.

\subsection{Feature Extraction and Classification Using Machine Learning}
\begin{figure}[htbp]
    \centering
    \includegraphics[bb=0.000000 0.000000 945.000000 390.000000, width=10cm]{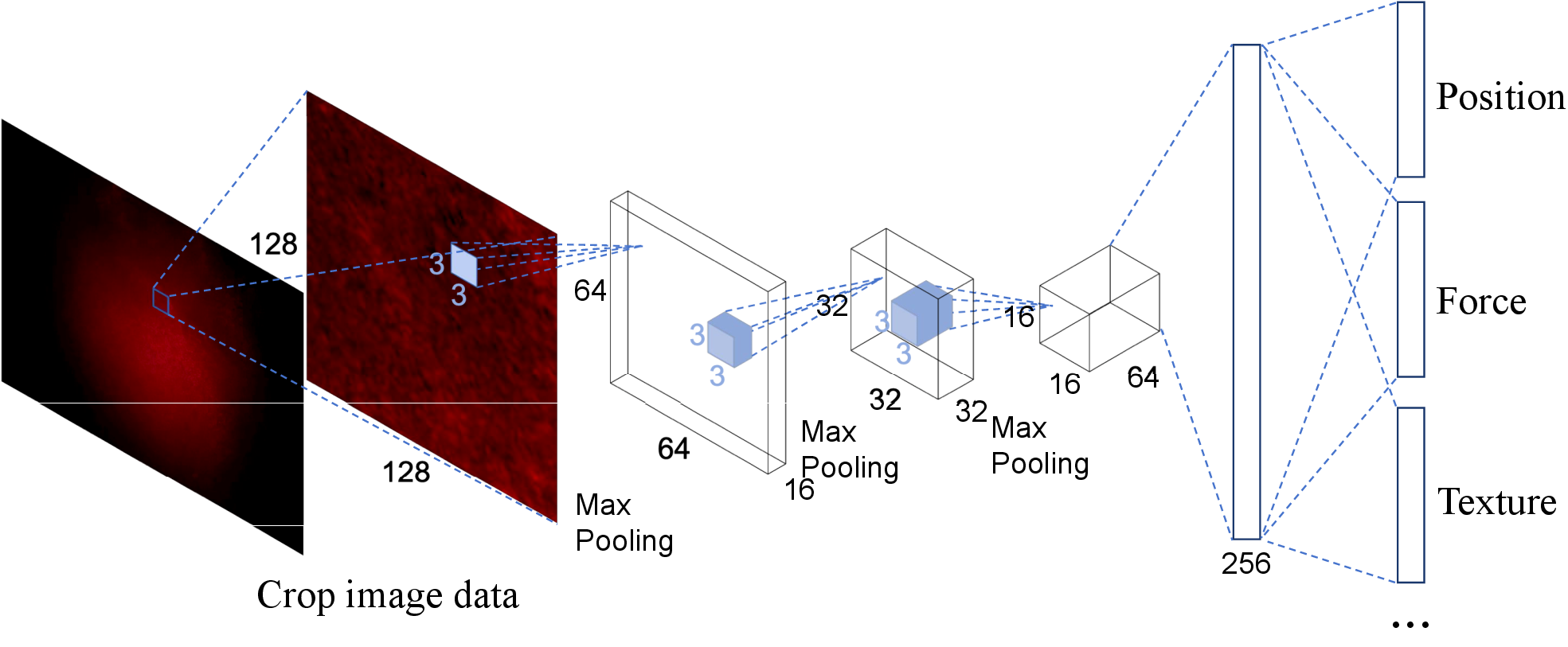}
    \caption{Decoding model based on CNN.}
\label{fig2}
\end{figure}

In this study, a lightweight convolutional neural network (CNN) was employed for the end-to-end classification of speckle patterns (Fig. ~\ref{fig2}). Prior to the network input, a central $128 \times 128$-pixel region was cropped from the raw image to focus on the region of deformation and reduce the computational load. The model processes single-channel input images and comprises three convolutional blocks followed by a fully connected classifier. Each block included a $3 \times 3$ convolutional layer, followed by batch normalization, a ReLU activation function, and $2 \times 2$ max pooling to reduce the spatial dimensions. After three downsampling operations, the feature maps are flattened and fed into a fully connected layer of 256 units, followed by batch normalization and ReLU activation. The final output layer decodes the sensing information, such as contact positions, contact force, or target classes, from the features. The network architecture was designed to trade-off between computational efficiency and classification accuracy. According to our estimation, the time of the inference was approximately 20.4 ms using a Raspberry Pi 5.

\section{Device design and fabrication}
The sensor was fabricated using a two-component liquid rubber (Verde Co., Ltd., Deco Resina\textregistered super clear silicone). The base compound and curing agent were thoroughly mixed at a 10:1 weight ratio and poured into an acrylic mold. Optionally, glass beads (diameter: 0.2 mm) were embedded in the mixture prior to curing. The mixture was left to cure at room temperature for 24-h until it was fully solidified.

After demolding, a small incision was performed at the edge of the silicone rubber using a blade. A polarization-maintaining(PM) fiber (Thorlabs, PM630-HP, flat cleaved, 1 m) was inserted through this opening to enable side-coupled laser illumination. After applying the mirror coating to the surface, a layer of silicone was applied to protect the surface. Simultaneously, a small amount of uncured liquid silicone was applied around the fiber entry point and camera (ArduCam, OV5647 Spy Camera Module for Raspberry Pi) mounting area to securely fix both the fiber and camera modules. This secondary bonding step ensured mechanical stability and minimized the impact of relative displacement. After an additional 24-h curing period, the fabrication process was complete, yielding a monolithic, flexible tactile sensor ready for operation.

\section{Experiments}
\subsection{Position recognition}
To evaluate the spatial perception ability of the proposed sensor, we experimentally assessed the ability of the proposed approach to recognize positions across arbitrary shapes.

\begin{figure}[htbp]
\centering\includegraphics[bb=0.000000 0.000000 356.000000 478.000000, width=10cm]{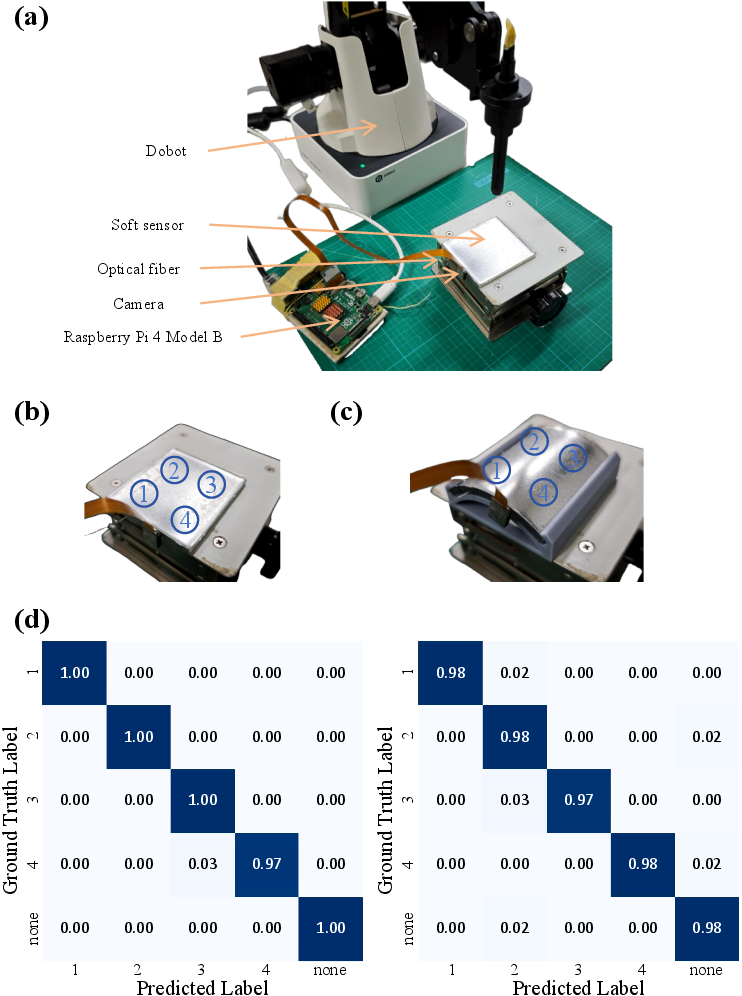}
\caption{(a) Experimental setup for position recognition. (b) Proposed sensor placement on a flat surface. (c) Proposed sensor placement on a curved surface. (d) Confusion matrix of the position recognition results. The left side shows the result placed on a flat surface; on the right is the result placed on a curved surface.}
\label{exp1}
\end{figure}

The experimental setup is shown in Fig.~\ref{exp1}(a). A rectangular silicone rubber $55 \times 61$ mm$^2$ was manufactured as the sensing element. The thickness was 3 mm. A PM fiber laser (Civil Laser, 635 nm, spectral linewidth: 1 nm) was used as the light source. The resulting speckle patterns were captured using a compact camera coupled with a Raspberry Pi 4 Model B. The sensor was tested in two configurations on a flat surface, as shown in Fig.~\ref{exp1}(b) and on a 3D-printed curved surface with a radius of curvature of 35 mm, as shown in Fig.~\ref{exp1}(c).

To collect training and test datasets, a robotic arm (DOBOT, Magician) was used to simulate a human finger pressing at four predefined positions across the sensor’s surface. Multiple force levels were applied at each position to investigate both the spatial location and robustness of the response. For each indentation event, a sequence of speckle images was recorded before and during contact.

The results of the position recognition on both flat and curved surfaces are presented in Fig.~\ref{exp1}(d). Note that the predicted contact locations closely align with the ground truth across all tested positions, demonstrating high spatial accuracy in both configurations. These results demonstrate that the proposed tactile sensor maintains a high performance in different working environments, confirming its shape-adaptive capability and robustness. 

\subsection{Force measurement}
\begin{figure}[htbp]
\centering\includegraphics[bb=0.000000 0.000000 367.000000 460.000000, width=12cm]{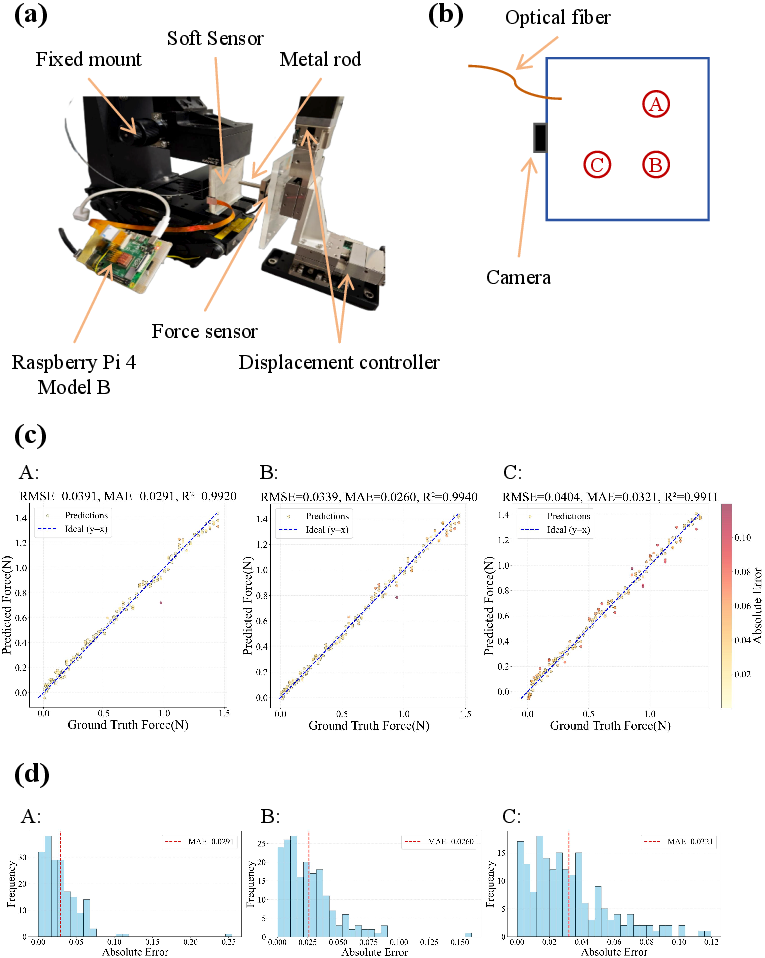}
\caption{(a) Experimental setup for force measurement. (b) Three locations (labeled “A”, “B”, and “C”) used for force measurement. (c) Scatter plots of the predicted forces versus the ground-truth values. RMSE and MAE denote the root-mean-square error and mean absolute error, respectively. $R^2$ is the coefficient of determination. (d) Histograms of the absolute errors.}
\label{exp2}
\end{figure}

We evaluated the ability of the proposed sensor to measure force. 
The experimental setup is shown in Fig.~\ref{exp2}(a). The left side shows a fixed mount, where the proposed sensor is vertically installed. The right side shows a displacement controller capable of providing precise motion in both the vertical and lateral directions. A force sensor (CM085-5N, Minebea Mitsumi Inc.) and a metal rod were attached to the controller, which was used to apply controlled indentation forces onto the sensor surface. This system enabled accurate and repeatable control of the contact force, allowing for consistent tactile stimulation during the experiments. 

Three different locations, labeled ``A'', ``B'', and ``C'', were selected to test the force measurement capability, as shown in Fig.~\ref{exp2}(b). The metal rod was driven by a displacement controller to apply controlled forces at the designated positions. After each displacement step, the force sensor readings and the corresponding speckle patterns were recorded. Each measurement was repeated three times to account for potential camera noise and ensure data consistency. The above procedure was repeated to collect training and test datasets.

The predicted forces were compared to the ground truth, as shown in Fig.~\ref{exp2}(c). A strong correlation between the predicted and actual forces was observed across all three locations. This result indicates a consistent performance regardless of the contact location. 
The quantitative evaluation yielded mean absolute errors (MAEs) of 0.0291\,N, 0.0260\,N, and 0.0321\,N, with corresponding root-mean-square errors (RMSEs) of 0.0391\,N, 0.0339\,N, and 0.0404\,N for the three tested locations. Figure~\ref{exp2}(d) shows the histograms of absolute errors. The errors exhibit a narrow distribution centered around zero, with most predictions falling within a small margin of error. The low error values and narrow variations across positions demonstrate high prediction accuracy and spatial uniformity in force estimation.

\subsection{Texture recognition}
\begin{figure}[htbp]
\centering\includegraphics[bb=0.000000 0.000000 325.000000 388.000000, width=10cm]{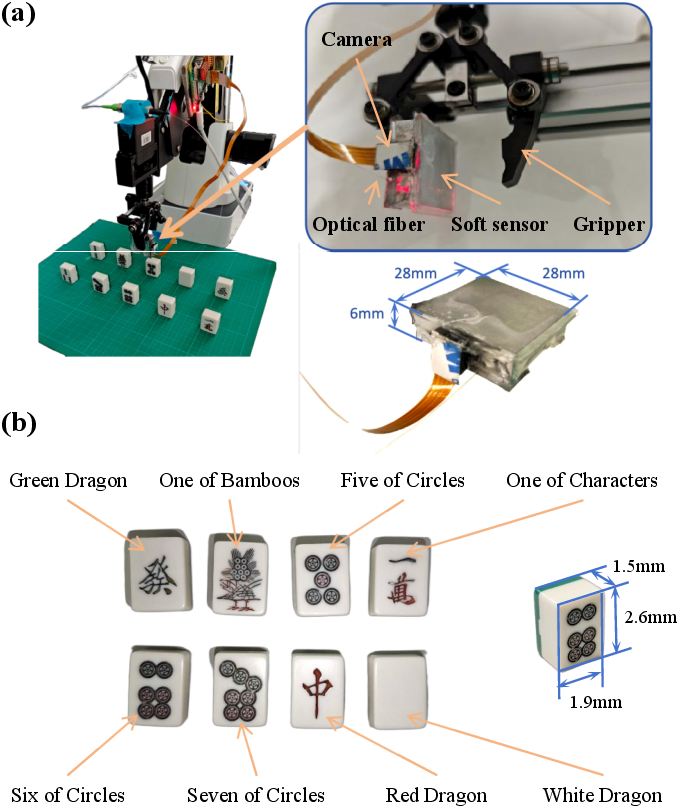}
\caption{(a) Experimental setup for Mahjong tile classification and the tactile sensor. The tactile sensor was designed to be mounted on a robotic gripper and used to grasp different tiles. (b) Mahjong tiles used in this experiment. The tiles are rigid cubes measuring $1.9,\text{mm} \times 1.5,\text{mm} \times 2.6,\text{mm}$, with different patterns engraved on their surfaces.}
\label{exp3_1}
\end{figure}

A more complex recognition task was conducted to demonstrate the texture recognition of the proposed tactile sensing approach. 
The experimental setup is shown in Fig.~\ref{exp3_1}(a). To integrate the sensor with the robotic gripper, a soft tactile module with dimensions 28 mm $\times$ 28 mm $\times$ 6 mm was fabricated using the same materials and manufacturing process [Fig.~\ref{exp3_1}(a)]. The sensor was mounted on the gripper and used to grasp various tiles, thereby enabling tactile interactions through controlled contact.

We utilized Mahjong tiles as the classification objects. Mahjong tiles (as shown in Fig.~\ref{exp3_1}(b)) are small rectangular game pieces with engraved surface patterns and varying textures, which are typically fabricated of plastic or resin. Mahjong tiles were used as textured objects with subtle surface variations. Nine distinct classes were considered in this experiment: eight Mahjong tiles---namely, ``White Dragon,'' ``Red Dragon,'' ``Green Dragon,'' ``One of Characters,'' ``One of Bamboos,'' ``Five of Circles,'' ``Six of Circles,'' and ``Seven of Circles''---as well as a ninth class representing the ``no-contact'' state. This expanded classification task tested the sensor's ability to distinguish subtle differences in surface patterns and geometries.
In total, 200 training and 40 test samples were collected from each class under consistent gripping conditions. The speckle patterns generated during contact were captured in real-time using a Raspberry Pi 4 Model B equipped with a camera module. 

\begin{figure}[htbp]
\centering\includegraphics[bb=0.000000 0.000000 326.000000 285.000000, width=10cm]{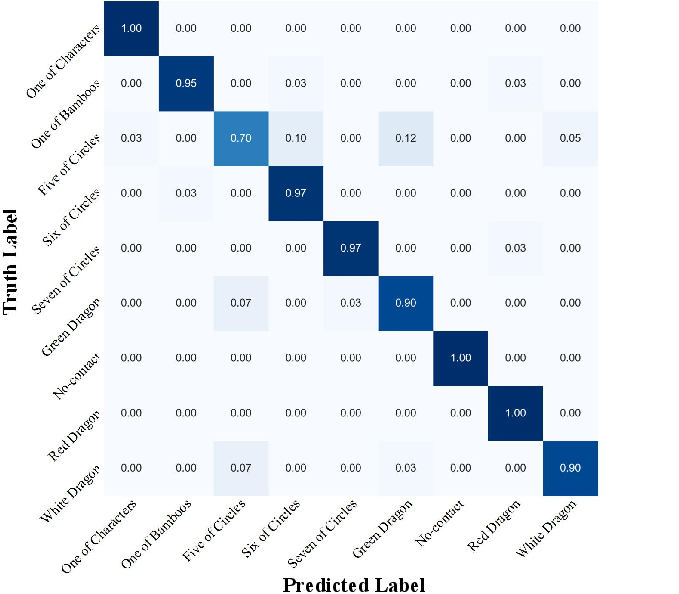}
\caption{Confusion matrix for the nine-class Mahjong tile classification task. The speckle images cropped from the region labeled “A” in Fig.~\ref{exp3_3}(a) were used for classification. The overall accuracy was 93.33\%.}
\label{exp3_2}
\end{figure}

\begin{figure}[htbp]
\centering\includegraphics[bb=0.000000 0.000000 325.000000 385.000000, width=9cm]{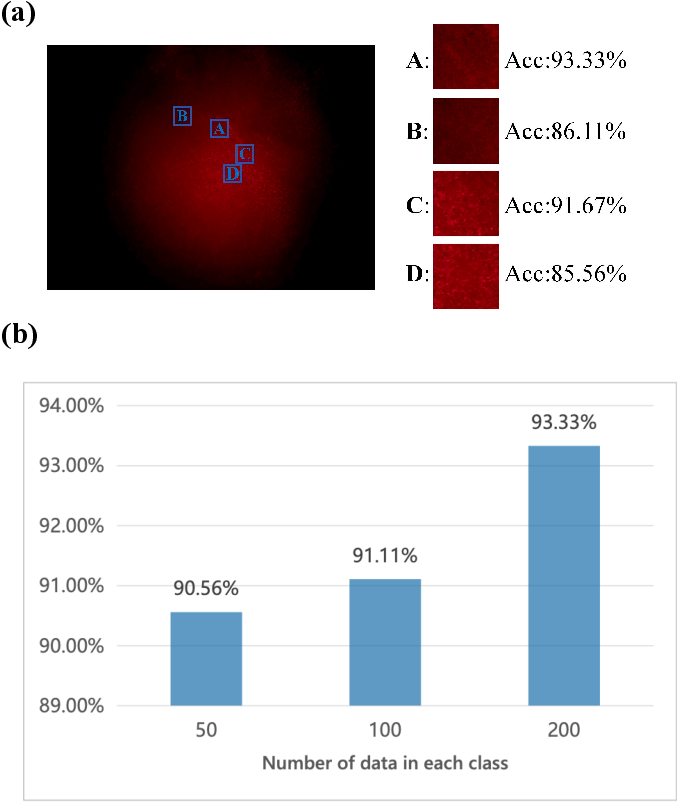}
\caption{(a) Cropped speckle images and the resulting classification accuracies. Four different image regions (labeled “A”, “B”, “C”, and “D”) were cropped from the original speckle image and used for the classification task. The classification accuracies were 93.33\%, 86.11\%, 91.67\%, and 85.56\%, respectively. (b) Classification accuracy as a function of the number of training samples per class.}
\label{exp3_3}
\end{figure}

Figure~\ref{exp3_2} shows the results of the nine-class recognition task. In this experiment, we used speckle images cropped from the region labeled “A” in Fig.~\ref{exp3_3}(a). The confusion matrix in Fig.~\ref{exp3_2} yielded an overall accuracy of 93.33\%, demonstrating the sensor’s ability to distinguish fine surface textures based on speckle pattern evolution.

To further investigate the classification capability, we also used speckle images cropped from different regions labeled “B,” “C,” and “D” (see Fig.~\ref{exp3_3}(a)). Classification accuracy varied depending on the selected region, achieving 86.11\%, 91.67\%, and 85.56\%, respectively. While all regions maintained acceptable performance, the results indicate that performance depends on the local light path density or proximity to the illumination edge.

Furthermore, to assess data efficiency, we reduced the training set size per class from 200 to 50 samples while keeping the test set constant. Interestingly, as shown in Fig.~\ref{exp3_3}(b), classification accuracy remained high at 90.56\%. This result highlights the effectiveness of the speckle-based representation in capturing discriminative tactile features even with limited training data.

\section{Discussion}
Table~\ref{comparison_table} presents a comparison between the proposed sensor and representative VBTSs. The proposed sensor offers several key advantages, including structural simplicity, a low profile, and shape adaptability.

Conventional VBTSs typically require many components, including an illumination unit, camera, lenses, mounting holders, housing, and a soft elastomer. The required components total at least seven, as shown in Table~\ref{comparison_table}. The lenses must be precisely focused on the surface of the soft elastomer to capture deformation-induced visual changes. This requirement necessitates rigid mechanical housings to maintain strict optical alignment and a fixed standoff distance between the lens and the sensing surface. Moreover, to enhance the visibility of subtle deformations, many designs incorporate elaborate internal structures, such as the dual-dome pigment layer in ChromaTouch~\cite{scharff2022rapid} or the multi-material textured tip in BioTacTip~\cite{BioTacTip2024}. These strategies are effective in improving signal contrast but significantly increase fabrication complexity and component count. Collectively, these factors lead to bulky, non-compliant systems that are ill-suited for integration into compact or highly flexible platforms.

In contrast, the proposed sensor was designed to operate without precise optical alignment and with a minimal number of components. This feature originates from the speckle-based sensing principle based on optical interference. While optical interferometric (speckle-based) approaches are highly sensitive, they also support large-area sensing through an alignment-free design and require only minimal core components: a laser source, a soft silicone medium, an optical fiber, and a compact camera. This simplified configuration significantly reduces fabrication complexity and overall system volume. Notably, the design does not require a large optical system to capture the full field of view of the sensor. As a result, the proposed sensor achieves a thin profile with an overall height of less than 3\,mm, which is an order of magnitude thinner than that of conventional systems.

Moreover, the monolithic and flexible architecture enables the sensor to conform to arbitrary surfaces, offering superior mechanical compliance compared to rigid counterparts.
In addition, the high sensitivity of the speckle pattern enabled precise force estimation, achieving a root-mean-square error (RMSE) of 0.04\,N (40\,mN), thereby demonstrating its capability for fine tactile perception.

\begin{table}
    \centering
    \caption{Comparison between the proposed sensor and recent representative VBTS approaches. Components indicate the minimal number of components to build the system(including light source, soft materials, optical system, and etc). Prediction error indicates the force measurement error. Flexibility indicates whether the system can be flexibly adapted to different shapes.}
    \label{comparison_table}
    \begin{tabular}{|c|c|c|c|}
        \hline
        Name & \textbf{This work} & ChromaTouch~\cite{scharff2022rapid} & GelSlim 3.0~\cite{gelslim3} \\
        \hline
        Principle & \textbf{Speckle based} & Color based & Colored shading based \\
        Overall Height(mm) & \textbf{3 $\sim$ 6} & - & 20 \\
        Required Resolution & \textbf{128$\times$128} & 1600$\times$1200 & 640$\times$480 \\
        Prediction Error(N) & \textbf{0.04} & - & - \\
        Components & \textbf{4} & 8 & 10 \\
	Optical Alignment & \textbf{Not required} & Required & Required \\
        Shape & Surface & Hemisphere & Flat \\
        Flexibility & \textbf{yes} & no & no \\
        \hline
        \hline
        Name &  BioTacTip~\cite{BioTacTip2024} & Digit360~\cite{lambeta2024digitizing} & ViTacTip~\cite{vitactip2024}\\
        \hline
        Principle & Marker based & Reflective layer based & Intensity based \\
        Overall Height(mm) & 45 & - & -\\
        Required Resolution & 1920$\times$1080 & - & 1920$\times$1080\\
        Prediction Error(N) & 0.1 & 0.001 & 0.04\\
        Components & 9 & 7 & 8\\
	Optical Alignment & Required & Required & Required \\
        Shape & Flat & Fingertip & Flat\\
        Flexibility & no & no & no\\
 \hline
    \end{tabular}
\end{table}

\section{Conclusion}
In this study, we demonstrated a highly sensitive, speckle-based optical soft tactile sensor featuring a simple structure, low profile, and shape-adaptive capability. Experimental results showed that the proposed sensor can accurately recognize both contact location and applied force with a precision of 40 mN. The high accuracy achieved in tasks such as Mahjong tile classification highlights the sensor’s sensitivity to fine surface features.

Future work includes optimizing the fabrication process, developing more efficient computational methods to improve system responsiveness, and exploring better integration strategies. These improvements will expand the applicability of the sensor to broader scenarios, including wearable devices, soft robotics, and human-machine interfaces.

\begin{acknowledgements}
This work was supported in part by 
the Japan Society for the Promotion of Science(JSPS) KAKENHI (Grant Nos. ~JP22K18792, JP22H05198, JP23K28157, JP25K22086),
and Japan Science and Technology Agency(JST), Core Research for Evolutional Science and Technology(CREST) (Grant No. ~JPMJCR24R2).
\end{acknowledgements}

\bibliography{references}

@article{Wang2015,
author = {Wang, Xiandi and Dong, Lin and Zhang, Hanlu and Yu, Ruomeng and Pan, Caofeng and Wang, Zhong Lin},
title = {Recent Progress in Electronic Skin},
journal = {Advanced Science},
volume = {2},
number = {10},
pages = {1500169},
doi = {https://doi.org/10.1002/advs.201500169},
url = {https://advanced.onlinelibrary.wiley.com/doi/abs/10.1002/advs.201500169},
year = {2015}
}

@article{Yang2019,
author = {Yang, Jun Chang and Mun, Jaewan and Kwon, Se Young and Park, Seongjun and Bao, Zhenan and Park, Steve},
title = {Electronic Skin: Recent Progress and Future Prospects for Skin-Attachable Devices for Health Monitoring, Robotics, and Prosthetics},
journal = {Advanced Materials},
volume = {31},
number = {48},
pages = {1904765},
year = {2019}
}

@article{MERIBOUT2024115332,
title = {Tactile sensors: A review},
journal = {Measurement},
volume = {238},
pages = {115332},
year = {2024},
issn = {0263-2241},
author = {Mahmoud Meribout and Natnael {Abule Takele} and Olyad Derege and Nidal Rifiki and Mohamed {El Khalil} and Varun Tiwari and Jing Zhong},
}

@article{Hammock2013,
author = {Hammock, Mallory L. and Chortos, Alex and Tee, Benjamin C.-K. and Tok, Jeffrey B.-H. and Bao, Zhenan},
title = {25th Anniversary Article: The Evolution of Electronic Skin (E-Skin): A Brief History, Design Considerations, and Recent Progress},
journal = {Advanced Materials},
volume = {25},
number = {42},
pages = {5997-6038},
year = {2013}
}

@article{Sekitani2012,
  author    = {Tsuyoshi, Sekitani and Takao, Someya},
  title     = {Stretchable organic integrated circuits for large-area electronic skin surfaces},
  journal   = {MRS Bulletin},
  year      = {2012},
  volume    = {37},
  number    = {3},
  pages     = {236--245},
  issn      = {1938-1425},
  date      = {2012-03-01},
}

@article{Jung2014,
author = {Jung, Sungmook and Kim, Ji Hoon and Kim, Jaemin and Choi, Suji and Lee, Jongsu and Park, Inhyuk and Hyeon, Taeghwan and Kim, Dae-Hyeong},
title ={Reverse-Micelle-Induced Porous Pressure-Sensitive Rubber for Wearable Human-Machine Interfaces},
journal = {Advanced Materials},
volume = {26},
number = {28},
pages = {4825-4830},
year = {2014}
}

@article{LEOGRANDE2025100312,
title = {Electronic skin technologies: From hardware building blocks and tactile sensing to control algorithms and applications},
journal = {Sensors and Actuators Reports},
volume = {9},
pages = {100312},
year = {2025},
issn = {2666-0539},
author = {Elisabetta Leogrande and Mariangela Filosa and Sara Ballanti and Luca {De Cicco} and Stefano Mazzoleni and Rochelle Ackerley and Calogero Maria Oddo and Francesco Dell'Olio},
}

@article{VEDEL1982289,
title = {Response to pressure and vibration of slowly adapting cutaneous mechanoreceptors in the human foot},
journal = {Neuroscience Letters},
volume = {34},
number = {3},
pages = {289-294},
year = {1982},
issn = {0304-3940},
author = {J.P. Vedel and J.P. Roll},
}

@article{clippinger1974,
  title={A sensory feedback system for an upper-limb amputation prosthesis},
  author={Clippinger, F W and Avery, R and Titus, B R},
  journal={Bulletin of Prosthetics Research},
  pages={247--258},
  year={1974},
}

@article{Grzegorz2005,
    author = {Darlinski, Grzegorz and Böttger, Ulrich and Waser, Rainer and Klauk, Hagen and Halik, Marcus and Zschieschang, Ute and Schmid, Günter and Dehm, Christine},
    title = {Mechanical force sensors using organic thin-film transistors},
    journal = {Journal of Applied Physics},
    volume = {97},
    number = {9},
    pages = {093708},
    year = {2005},
    month = {04},
    issn = {0021-8979},
}

@Inbook{Cheng2009,
author="Cheng, I-Chun and Wagner, Sigurd and Wong, William S. and Salleo, Alberto",
title="Overview of Flexible Electronics Technology",
bookTitle="Flexible Electronics: Materials and Applications",
year="2009",
publisher="Springer US",
address="Boston, MA",
pages="1--28",
isbn="978-0-387-74363-9",
}

@Article{Stassi2014,
AUTHOR = {Stassi, Stefano and Cauda, Valentina and Canavese, Giancarlo and Pirri, Candido Fabrizio},
TITLE = {Flexible Tactile Sensing Based on Piezoresistive Composites: A Review},
JOURNAL = {Sensors},
VOLUME = {14},
YEAR = {2014},
NUMBER = {3},
PAGES = {5296--5332},
URL = {https://www.mdpi.com/1424-8220/14/3/5296},
PubMedID = {24638126},
ISSN = {1424-8220},
}

@article{Choong2014,
author = {Choong, Chwee-Lin and Shim, Mun-Bo and Lee, Byoung-Sun and Jeon, Sanghun and Ko, Dong-Su and Kang, Tae-Hyung and Bae, Jihyun and Lee, Sung Hoon and Byun, Kyung-Eun and Im, Jungkyun and Jeong, Yong Jin and Park, Chan Eon and Park, Jong-Jin and Chung, U-In},
title = {Highly Stretchable Resistive Pressure Sensors Using a Conductive Elastomeric Composite on a Micropyramid Array},
journal = {Advanced Materials},
volume = {26},
number = {21},
pages = {3451-3458},
year = {2014}
}

@INPROCEEDINGS{6584238,
  author={Kõiva, Risto and Zenker, Matthias and Schürmann, Carsten and Haschke, Robert and Ritter, Helge J.},
  booktitle={2013 IEEE/ASME International Conference on Advanced Intelligent Mechatronics}, 
  title={A highly sensitive 3D-shaped tactile sensor}, 
  year={2013},
  volume={},
  number={},
  pages={1084-1089},
  doi={10.1109/AIM.2013.6584238}}

@INPROCEEDINGS{1641896,
  author={Ohmura, Y. and Kuniyoshi, Y. and Nagakubo, A.},
  booktitle={Proceedings 2006 IEEE International Conference on Robotics and Automation, 2006. ICRA 2006.}, 
  title={Conformable and scalable tactile sensor skin for curved surfaces}, 
  year={2006},
  volume={},
  number={},
  pages={1348-1353},
  doi={10.1109/ROBOT.2006.1641896}}

@article{Qi2023,
author = {Qi, FangXi and Xu, Lei and He, Yin and Yan, Han and Liu, Hao},
title = {PVDF-Based Flexible Piezoelectric Tactile Sensors: Review},
journal = {Crystal Research and Technology},
volume = {58},
number = {10},
pages = {2300119},
doi = {https://doi.org/10.1002/crat.202300119},
url = {https://onlinelibrary.wiley.com/doi/abs/10.1002/crat.202300119},
year = {2023}
}

@article{RYU2022131308,
title = {PVDF-bismuth titanate based self-powered flexible tactile sensor for biomechanical applications},
journal = {Materials Letters},
volume = {309},
pages = {131308},
year = {2022},
issn = {0167-577X},
doi = {https://doi.org/10.1016/j.matlet.2021.131308},
url = {https://www.sciencedirect.com/science/article/pii/S0167577X21020061},
author = {Chaehyun Ryu and Sugato Hajra and Manisha Sahu and Soon In Jung and Il Ryu Jang and Hoe Joon Kim},
}

@article{Lee2011,
doi = {10.1088/0960-1317/21/3/035010},
url = {https://dx.doi.org/10.1088/0960-1317/21/3/035010},
year = {2011},
month = {feb},
publisher = {},
volume = {21},
number = {3},
pages = {035010},
author = {Lee, Hyung-Kew and Chung, Jaehoon and Chang, Sun-Il and Yoon, Euisik},
title = {Real-time measurement of the three-axis contact force distribution using a flexible capacitive polymer tactile sensor},
journal = {Journal of Micromechanics and Microengineering},
}

@article{Lipomi2011,
  author    = {Lipomi, Darren J. and Vosgueritchian, Michael and Tee, Benjamin C-K. and Hellstrom, Sondra L. and Lee, Jennifer A. and Fox, Courtney H. and Bao, Zhenan},
  title     = {Skin-like pressure and strain sensors based on transparent elastic films of carbon nanotubes},
  journal   = {Nature Nanotechnology},
  year      = {2011},
  volume    = {6},
  number    = {12},
  pages     = {788--792},
  issn      = {1748-3395},
  date      = {2011-12-01},
}

@article{Zang2015,
  author    = {Zang, Yaping and Zhang, Fengjiao and Huang, Dazhen and Gao, Xike and Di, Chong-an and Zhu, Daoben},
  title     = {Flexible suspended gate organic thin-film transistors for ultra-sensitive pressure detection},
  journal   = {Nature Communications},
  year      = {2015},
  volume    = {6},
  number    = {1},
  pages     = {6269},
  issn      = {2041-1723},
  date      = {2015-03-03},
}

@article{YAO2024111040,
title = {Recent progress of optical tactile sensors: A review},
journal = {Optics and Laser Technology},
volume = {176},
pages = {111040},
year = {2024},
issn = {0030-3992},
author = {Ni Yao and Shipeng Wang},
}

@article{Mohamed2022,
author = {Elsherif, Mohamed and Salih, Ahmed E. and Muñoz, Monserrat Gutiérrez and Alam, Fahad and AlQattan, Bader and Antonysamy, Dennyson Savariraj and Zaki, Mohamed Fawzi and Yetisen, Ali K. and Park, Seongjun and Wilkinson, Timothy D. and Butt, Haider},
title = {Optical Fiber Sensors: Working Principle, Applications, and Limitations},
journal = {Advanced Photonics Research},
volume = {3},
number = {11},
pages = {2100371},
year = {2022}
}

@article{Kao1966,
author = {K.C. Kao  and G.A. Hockham },
title = {Dielectric-fibre surface waveguides for optical frequencies},
journal = {Proceedings of the Institution of Electrical Engineers},
volume = {113},
issue = {7},
pages = {1151-1158},
year = {1966},
}

@Article{s17010111,
AUTHOR = {Fajkus, Marcel and Nedoma, Jan and Martinek, Radek and Vasinek, Vladimir and Nazeran, Homer and Siska, Petr},
TITLE = {A Non-Invasive Multichannel Hybrid Fiber-Optic Sensor System for Vital Sign Monitoring},
JOURNAL = {Sensors},
VOLUME = {17},
YEAR = {2017},
NUMBER = {1},
ARTICLE-NUMBER = {111},
ISSN = {1424-8220},
DOI = {10.3390/s17010111}
}

@article{Mizuno:21,
author = {Yosuke Mizuno and Antreas Theodosiou and Kyriacos Kalli and Sascha Liehr and Heeyoung Lee and Kentaro Nakamura},
journal = {Photon. Res.},
number = {9},
pages = {1719--1733},
publisher = {Optica Publishing Group},
title = {Distributed polymer optical fiber sensors: a review and outlook},
volume = {9},
month = {Sep},
year = {2021},
}

@article{Ma2002,
author = {Ma, H. and Jen, A.K.-Y. and Dalton, L.R.},
title = {Polymer-Based Optical Waveguides: Materials, Processing, and Devices},
journal = {Advanced Materials},
volume = {14},
number = {19},
pages = {1339-1365},
year = {2002}
}

@article{Guo2016,
  title={Highly stretchable, strain sensing hydrogel optical fibers},
  author={Guo, Jingjing and Liu, Xinyue and Jiang, Nan and Yetisen, Ali K and Yuk, Hyunwoo and Yang, Changxi and Khademhosseini, Ali and Zhao, Xuanhe and Yun, Seok-Hyun},
  journal={Advanced Materials (Deerfield Beach, Fla.)},
  volume={28},
  number={46},
  pages={10244},
  year={2016}
}

@article{Wang2020Hydrogel,
author = {Wang, Chengmin and Wu, Baohu and Sun, Shengtong and Wu, Peiyi},
title = {Interface Deformable, Thermally Sensitive Hydrogel–Elastomer Hybrid Fiber for Versatile Underwater Sensing},
journal = {Advanced Materials Technologies},
volume = {5},
number = {12},
pages = {2000515},
year = {2020}
}

@article{ELSHERIF201925,
title = {Hydrogel optical fibers for continuous glucose monitoring},
journal = {Biosensors and Bioelectronics},
volume = {137},
pages = {25-32},
year = {2019},
issn = {0956-5663},
author = {Mohamed Elsherif and Muhammad Umair Hassan and Ali K. Yetisen and Haider Butt},
}

@article{zhang2020mnf,
  title={Ultrasensitive skin-like wearable optical sensors based on glass micro/nanofibers},
  author={Zhang, Lei and Pan, Jing and Zhang, Zhang and Wu, Hao and Yao, Ni and Cai, Dawei and Xu, Yingxin and Zhang, Jin and Sun, Guofei and Wang, Liqiang and others},
  journal={Opto-Electronic Advances},
  volume={3},
  number={3},
  pages={190022--1},
  year={2020},
  publisher={Opto-Electronic Advances}
}

@article{Jiang2021,
author = {Jiang, Chengpeng and Zhang, Zhang and Pan, Jing and Wang, Yancheng and Zhang, Lei and Tong, Liming},
title = {Finger-Skin-Inspired Flexible Optical Sensor for Force Sensing and Slip Detection in Robotic Grasping},
journal = {Advanced Materials Technologies},
volume = {6},
number = {10},
pages = {2100285},
year = {2021}
}

@Article{Pan2020,
author ={Pan, Jing and Zhang, Zhang and Jiang, Chengpeng and Zhang, Lei and Tong, Limin},
title  ={A multifunctional skin-like wearable optical sensor based on an optical micro-/nanofibre},
journal  ={anoscale},
year  ={2020},
volume  ={12},
issue  ={33},
pages  ={17538-17544},
publisher  ={The Royal Society of Chemistry}
}

@article{VBTSreviewAPR,
    author = {Xin, Yi-Hang and Hu, Kai-Ming and Xiang, Rui-Jia and Gao, Yu-Ling and Zhou, Jun-Feng and Meng, Guang and Zhang, Wen-Ming},
    title = {Vision-based tactile sensing: From performance parameters to device design},
    journal = {Applied Physics Reviews},
    volume = {12},
    number = {2},
    pages = {021312},
    year = {2025},
    month = {04},
}

@ARTICLE{VBTSreviewIEEE,
  author={Li, Haoran and Lin, Yijiong and Lu, Chenghua and Yang, Max and Psomopoulou, Efi and Lepora, Nathan F.},
  journal={IEEE Sensors Journal}, 
  title={Classification of Vision-Based Tactile Sensors: A Review}, 
  year={2025},
  volume={25},
  number={19},
  pages={35672-35686},
  doi={10.1109/JSEN.2025.3599236}}

@ARTICLE{GelForce,
  author={Sato, Katsunari and Kamiyama, Kazuto and Kawakami, Naoki and Tachi, Susumu},
  journal={IEEE Transactions on Haptics}, 
  title={Finger-Shaped GelForce: Sensor for Measuring Surface Traction Fields for Robotic Hand}, 
  year={2010},
  volume={3},
  number={1},
  pages={37-47},
  doi={10.1109/TOH.2009.47}}

@Article{gelsight,
AUTHOR = {Yuan, Wenzhen and Dong, Siyuan and Adelson, Edward H.},
TITLE = {GelSight: High-Resolution Robot Tactile Sensors for Estimating Geometry and Force},
JOURNAL = {Sensors},
VOLUME = {17},
YEAR = {2017},
NUMBER = {12},
ARTICLE-NUMBER = {2762},
URL = {https://www.mdpi.com/1424-8220/17/12/2762},
PubMedID = {29186053},
ISSN = {1424-8220},
}

@INPROCEEDINGS{gelslim,
  author={Donlon, Elliott and Dong, Siyuan and Liu, Melody and Li, Jianhua and Adelson, Edward and Rodriguez, Alberto},
  booktitle={2018 IEEE/RSJ International Conference on Intelligent Robots and Systems (IROS)}, 
  title={GelSlim: A High-Resolution, Compact, Robust, and Calibrated Tactile-sensing Finger}, 
  year={2018},
  volume={},
  number={},
  pages={1927-1934},
  }

@INPROCEEDINGS{gelslim3,
  author={Taylor, Ian H. and Dong, Siyuan and Rodriguez, Alberto},
  booktitle={2022 International Conference on Robotics and Automation (ICRA)}, 
  title={GelSlim 3.0: High-Resolution Measurement of Shape, Force and Slip in a Compact Tactile-Sensing Finger}, 
  year={2022},
  volume={},
  number={},
  pages={10781-10787},
  }

@article{lambeta2024digitizing,
  title={Digitizing touch with an artificial multimodal fingertip},
  author={Lambeta, Mike and Wu, Tingfan and Sengul, Ali and Most, Victoria Rose and Black, Nolan and Sawyer, Kevin and Mercado, Romeo and Qi, Haozhi and Sohn, Alexander and Taylor, Byron and others},
  journal={arXiv preprint arXiv:2411.02479},
  year={2024}
}

@article{Shimadera2022,
  author    = {Shimadera, Sho and Kitagawa, Kei and Sagehashi, Koyo and Miyajima, Yoji and Niiyama, Tomoaki and Sunada, Satoshi},
  title     = {Speckle-based high-resolution multimodal soft sensing},
  journal   = {Scientific Reports},
  year      = {2022},
  volume    = {12},
  number    = {1},
  pages     = {13096},
  issn      = {2045-2322},
  date      = {2022-07-30},
}

@article{Kitagawa2024,
author = {Kei Kitagawa and Kohei Tsuji and Koyo Sagehashi and Tomoaki Niiyama and Satoshi Sunada},
journal = {Opt. Express},
number = {3},
pages = {3209--3220},
publisher = {Optica Publishing Group},
title = {Optical hyperdimensional soft sensing: speckle-based touch interface and tactile sensor},
volume = {32},
month = {Jan},
year = {2024},
}

@inproceedings{scharff2022rapid,
  title={Rapid manufacturing of color-based hemispherical soft tactile fingertips},
  author={Scharff, Rob BN and Boonstra, Dirk-Jan and Willemet, Laurence and Lin, Xi and Wiertlewski, Micha{\"e}l},
  booktitle={2022 IEEE 5th international conference on soft robotics (RoboSoft)},
  pages={896--902},
  year={2022},
  organization={IEEE}
}

@ARTICLE{BioTacTip2024,
  author={Li, Haoran and Nam, Saekwang and Lu, Zhenyu and Yang, Chenguang and Psomopoulou, Efi and Lepora, Nathan F.},
  journal={IEEE Robotics and Automation Letters}, 
  title={BioTacTip: A Soft Biomimetic Optical Tactile Sensor for Efficient 3D Contact Localization and 3D Force Estimation}, 
  year={2024},
  volume={9},
  number={6},
  pages={5314-5321},
  doi={10.1109/LRA.2024.3387111}}

@inproceedings{magictac2024,
  title={Magictac: A novel high-resolution 3d multi-layer grid-based tactile sensor},
  author={Fan, Wen and Li, Haoran and Zhang, Dandan},
  booktitle={2024 IEEE International Conference on Robotics and Automation (ICRA)},
  pages={388--394},
  year={2024},
  organization={IEEE}
}

@inproceedings{vitactip2024,
  title={Vitactip: Design and verification of a novel biomimetic physical vision-tactile fusion sensor},
  author={Fan, Wen and Li, Haoran and Si, Weiyong and Luo, Shan and Lepora, Nathan and Zhang, Dandan},
  booktitle={2024 IEEE International Conference on Robotics and Automation (ICRA)},
  pages={1056--1062},
  year={2024},
  organization={IEEE}
}

@Article{Csight2024,
AUTHOR = {Fan, Wen and Li, Haoran and Xing, Yifan and Zhang, Dandan},
TITLE = {Design and Evaluation of a Rapid Monolithic Manufacturing Technique for a Novel Vision-Based Tactile Sensor: C-Sight},
JOURNAL = {Sensors},
VOLUME = {24},
YEAR = {2024},
NUMBER = {14},
ARTICLE-NUMBER = {4603},
URL = {https://www.mdpi.com/1424-8220/24/14/4603},
PubMedID = {39066001},
ISSN = {1424-8220},
}

\end{document}